\pdfoutput=1

\documentclass[11pt]{article}

\usepackage[]{EMNLP2023}
\usepackage{enumitem}
\usepackage{times}

\usepackage{latexsym}
\usepackage{graphicx}
\usepackage{booktabs}

\usepackage[T1]{fontenc}

\usepackage[utf8]{inputenc}

\usepackage{microtype}

\usepackage{inconsolata}
\usepackage{colortbl} 

\usepackage{amsmath}
\title{Can Language capture Cognitive Dissonance?}
\title{\textit{I want to but I can't}: Can Language capture Cognitive Coherence in Decision-Making?}
\title{\textit{I want to but I can't}: Validating Cognitive Processes in Language}
\title{\texttt{CoPLa: }Validating Cognitive Processes with Language \\ Cognitive Coherence in Decision-Making}
\title{\textit{I want to but I can't}: Validating Cognitive Coherence with Language}
\title{ Can Language capture Cognitive Processes?\\ The Case of Decision-Making}

\title{\textit{I want to but I can't}: Towards Capturing Cognitive Styles with \\Language in the context of Decision Making}
\title{Capturing Human Cognitive Styles with Language: \\Towards an Experimental Evaluation Paradigm 
}

\author{
Vasudha Varadarajan$^\dagger$, Syeda Mahwish$^\dagger$, Xiaoran Liu$^\spadesuit$, Julia Buffolino$^\dagger$\\
\textbf{Christian C. Luhmann$^\spadesuit$, Ryan L. Boyd$^\clubsuit$, H. Andrew Schwartz$^\dagger$} \\
$^\dagger$Department of Computer Science, Stony Brook University \\
$^\spadesuit$Department of Psychology, Stony Brook University \\
$^\clubsuit$Department of Psychology, University of Texas at Dallas \\
{\tt  \{vvaradarajan, has\}@cs.stonybrook.edu} 
}

\begin{document}
\maketitle
\begin{abstract}

While NLP models often seek to capture cognitive states via language, the validity of predicted states is determined by comparing them to annotations created without access the cognitive states of the authors.
In behavioral sciences, cognitive states are instead measured via experiments. 
Here, we introduce an experiment-based framework 
for evaluating language-based cognitive style models against human behavior. 
We explore the phenomenon of decision making, and its relationship to the linguistic style of an individual talking about a recent decision they made. 
The participants then follow a classical decision-making experiment that captures their cognitive style, determined by how preferences change during a decision exercise. 
We find that language features, intended to capture cognitive style, can 
predict participants' decision style with moderate-to-high accuracy (AUC $\sim$ 0.8), demonstrating that cognitive style can be partly captured and revealed by discourse patterns.

\end{abstract}

\section{Introduction}

\begin{figure}[!ht]
 \centering
 \includegraphics[width=0.9\columnwidth]{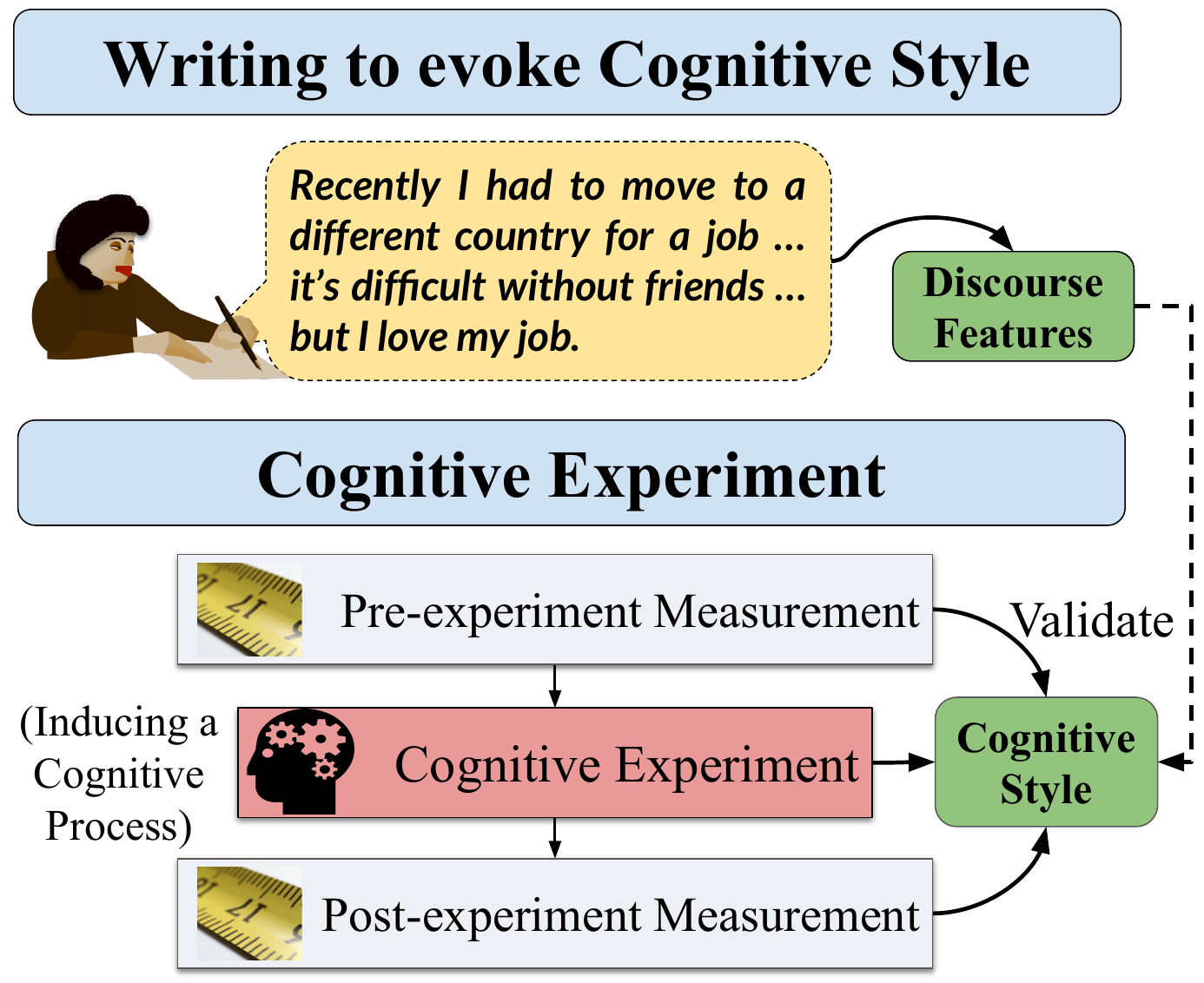}
 \caption{An alternate evaluation framework for validation of cognitive processes with language: The participants are first prompted to write about their experiences, eliciting their thought process. Then they are subjected to an experiment that would measure their behavior. The behavior is a \textit{ground truth} measure of their cognitive style that can be tied to the expressed language. 
 }
 \label{fig:spirit}
\end{figure}

While language models grow in sophistication, NLP tasks increasingly focus on understanding the people behind the language~\cite{choi-etal-2023-llms, dey2024socialite}.
Such social and psychological NLP studies still rely primarily on \textbf{annotations} for evaluation. 
For example, recent social tasks have depended on annotated datasets for, e.g.,  emotions~\cite{rosenthal2019semeval,mohammad2018semeval}, empathy~\cite{sharma2020computational}, politeness~\cite{hayati-etal-2021-bert}, humor~\cite{meaney2021semeval}, dissonance~\cite{varadarajan-etal-2023}, and reasoning abilities~\cite{alhamzeh2022s}. 
However, while annotation-based work has pushed NLP towards capturing cognitive states of the language generators (i.e. people),
it falls short of offering \textit{ground truth} of psychological processes 
because annotations reflect \textit{perception} of another person's state.~\cite{sandri-etal-2023-dont, sap2021annotators}.
For example, annotations of empathy point to linguistic cues that \textit{appear} empathetic to observers but do not always reflect the actual human experience of empathy~\cite{lahnala-etal-2022-critical}. 
Behavioral sciences, on the other hand, often emphasize the importance of \textit{direct} assessment through experimental paradigms for the purpose of understanding constructs of interest.

We introduce an experimental framework that collects linguistic data alongside induced cognitive phenomena to evaluate the feasibility of discourse modeling approaches for capturing cognitive styles in decision making. By associating linguistic patterns with specific cognitive phenomena, we aim to understand individuals' unique cognitive styles, which are largely unseen and often only observable through the final decision~\cite{campitelli2010herbert}. Our study follows modern psychology--experimental designs to quantify how language use signals Cognitive Styles, or habitual patterns of thought related to various cognitive phenomena. 

Our key contributions include: (1) An experiment-based evaluation framework to validate cognitive styles in language;
(2) Exploring discourse and other linguistic features for modeling decision-making cognitive styles;
(3) Finding that language can be indicative of a person's cognitive styles even in the more stringent evaluation framework;
(4) \textit{Decisions} dataset for the language of decision-making cognitive styles.\footnote{For dataset and code: \hyperlink{https://github.com/humanlab/cog_style_validation}{https://github.com/humanlab/cog\_style\_validation}}





\section{Related Work}
While NLP for social science often relies on labels from annotators or questionnaires, behavioral science theory suggests carefully designed experiments can more objectively elicit and capture cognitive states.
For instance, \citet{saxbe2013embodiment} investigated emotional responses using an experimental design in which participants' brain activity was imaged as they listened to narratives eliciting different emotions. These methods can offer a more objective foundation for understanding the psychological processes at play~\citep{brook2015linking}. In our study, we focus on cognitive styles in decision making -- reflecting one's tendency to maintain consistency and resolve dissonance~\citep{harmon2007cognitive,mcgrath2017dealing}. 
Since people show little awareness of their decision-making~\cite{nisbett1977halo}, 
 cognitive styles are measured experimentally by observing shifts in preferences after decision~\citep{simon2004construction,aguilar2022cognitive}. 
Our study draws from previous NLP research that validates author state measurements through annotations or self-report questionnaires. For example, past work has compared affective states with self-reported mental health by analyzing self-disclosures~\citep{zirikly-etal-2019-clpsych,valizadeh-etal-2021-identifying}, while others have examined cognitive styles in the context of discourse~\citep{sharma-etal-2023-cognitive, juhng2023discourse, varadarajan2022disso,varadarajan-etal-2023}. However, annotations and self-reports are subject to perceptual biases, such as those observed in dialogue evaluations~\citep{liang2020beyond} or when assessing constructs such as humor, empathy, or offensiveness~\citep{yang2021choral, paulhus2007self,buechel-etal-2018-modeling, lahnala-etal-2024-appraisal-framework}. To address these limitations, we adopt an experimental approach that aims to objectively capture cognitive states, focusing on how individuals manage dissonance and consistency in their decision-making.

Discourse structures provide a theoretically grounded link between cognitive processes and communication patterns, serving as a window into how individuals construct and convey explanations~\cite{van1990social, van2014discourse}.
Research in psychology has established strong connections between linguistic patterns and cognitive styles, particularly in how individuals process and communicate information~\cite{buchanan2013explanatory}. The analysis of discourse relations is especially valuable because they capture both explicit and implicit connections between text segments, revealing deeper patterns in explanatory styles such as reasoning~\cite{son-etal-2017-recognizing,son-etal-2018-causal} and rhetorical structures~\citep{taboada2006rhetorical} that may not be apparent from lexical-level features alone~\cite{juhng2023discourse, varadarajan-etal-2024-archetypes}. It serves as a powerful indicator of explanatory and rhetorical patterns in text, offering insights into how ideas are connected and presented~\cite{knaebel2023discourse}. In this work, we explore discourse features as well as state-of-the-art LLMs to model the outcomes of the cognitive experiment.
\section{Experiment}

A total of 514 participants were recruited in person for the study; 12 were excluded due to incomplete or invalid responses, resulting in a final dataset of 502 participants. Data collection was performed in 2 stages (see Figure~\ref{fig:spirit}). The questionnaire has been described in detail in Appendix~\ref{sec:joboffer_desc}.
 
 \paragraph{Writing Task } Participants received 2 writing prompts to elicit language relevant to their decision-making cognitive style:
 1) ``Please describe a recent important and difficult decision that you have made'' (20-100 words), and 
 2) ``What were the considerations that you thought about while making the decision? When answering, please consider all of the circumstances and details that went into the difficult decision'' (100-300 words).
These questions were chosen to elicit detailed descriptions of a recent decision-making process, encouraging participants to discuss options and explain their reasoning. 
The elicited essays to the two questions were concatenated for all further analysis.
We henceforth call the  collection of essays from the participants the \textit{Decisions} dataset.

 \paragraph{Constraint Satisfaction Experiment}

 \label{subsec:job_offer}
We replicated the experiment from \citet{simon2004construction}, modifying the preference score calculation to quantify overall preference changes rather than single attribute fluctuations as described below.
 \paragraph{1. Pre-Decision Preferences} 
Participants answered questions assessing preferences on a 6-point scale (-5 to 5, interval of 2) for four attributes: 
 \newcommand{\f}{\mkern-2mu f\mkern-3mu}
 Commute ($com$), Vacation ($vac$), Office space ($o\f\f$) and Salary ($sal$).  
 
 Each attribute had positive (+) and negative (-) questions for preference bounds. 
 For example: 
 $com_{+}$ (commute): ``
 Please select how desirable the 18 minute commute is to you.'' 
 $com_{-}$: ``
 Please select how desirable the 40 minute commute is to you.'' 
 Participants rated each attribute's relative weight ($\texttt{W}$) on a 1-8 scale. 
 Final preference ($\rho_{\scriptsize\texttt{com}}$) for each attribute was calculated as:\\
\hspace*{2em} $\rho_{\scriptsize\texttt{com}} = (com^{+} - com^{-})\ \texttt{x}\ \texttt{W}_{com}$
\\Note that each $\texttt{$\rho$}$ is a value between -80 to +80.
\paragraph{2. Job Offers}
Two choices were offered to the participants, such that in choosing either of the jobs, they would likely make compromises on at least two attributes. The two options were: \\
Company A: $com_{+}$, $vac_{+}$, $o{\f\f}_{-}$, $sal_{-}$; and\\ Company B: $com_{-}$, $vac_{-}$, $o{\f\f}_{+}$, $sal_{+}$ \\where $_{+},_{-}$ (in subscript) refer to the favorable and unfavorable conditions for each of the four attributes.
Therefore, the pre-decision preference score $\psi$ for company A and B can be calculated as:
 $   \texttt{$\psi$}_{\texttt{A}}  = + \texttt{$\rho$}_{com} + \texttt{$\rho$}_{vac} -\texttt{$\rho$}_{o\f\f} - \texttt{$\rho$}_{sal} $\\
 $\texttt{$\psi$}_{\texttt{B}} = -\texttt{$\rho$}_{com} - \texttt{$\rho$}_{vac} +\texttt{$\rho$}_{o\f\f} + \texttt{$\rho$}_{sal} $\\
Each $\texttt{$\psi$}$ thus has a value between -320 and +320. Participants choose between two options, typically aligning with their initial preferences. 
We randomly introduce an influencing factor, location ($loc$), describing the job as either near a fun mall or in a dull construction site. This aims to induce dissonance, compelling participants to compromise and potentially make contrarian decisions, inconsistent with their initial preferences.
\paragraph{3. Post-Decision Preferences}
After selecting a job, participants answer the same questions from the pre-decision questionnaire again.

\paragraph{Decision-Making Outcomes}
We define each construct and describe their measurement from the experiment below:

\label{subsec:dm_outcomes}
\paragraph{1. Choice-Induced Shift (CIS)} The change in preference is captured by subtracting the pre-experiment scores from post-experiment scores.
$CIS = \texttt{$\psi$}^{post}_\texttt{A} - \texttt{$\psi$}^{pre}_\texttt{A}; \texttt{choice = A}$\\
$CIS = \texttt{$\psi$}^{post}_\texttt{B} - \texttt{$\psi$}^{pre}_\texttt{B};  \texttt{choice = B}$\\
Here, we model binarized CIS which captures the direction of the preference change towards the job choice.

\paragraph{2. Influenced or Not (Inf)} 
The job offer is further influenced by introducing a confounding attribute $loc$ (location). Many participants choose the job influenced by the description of the location attribute, however not all of them change their minds. The change is a cognitive signal that measures if someone's choice was influenced by confounding attribute. This indicates that their initial preferences were not strong enough to begin with. This is captured as a binary variable: making the choice in the direction of the influenced variable or not.

\begin{figure}[ht]
 \centering
 \includegraphics[width=\linewidth]{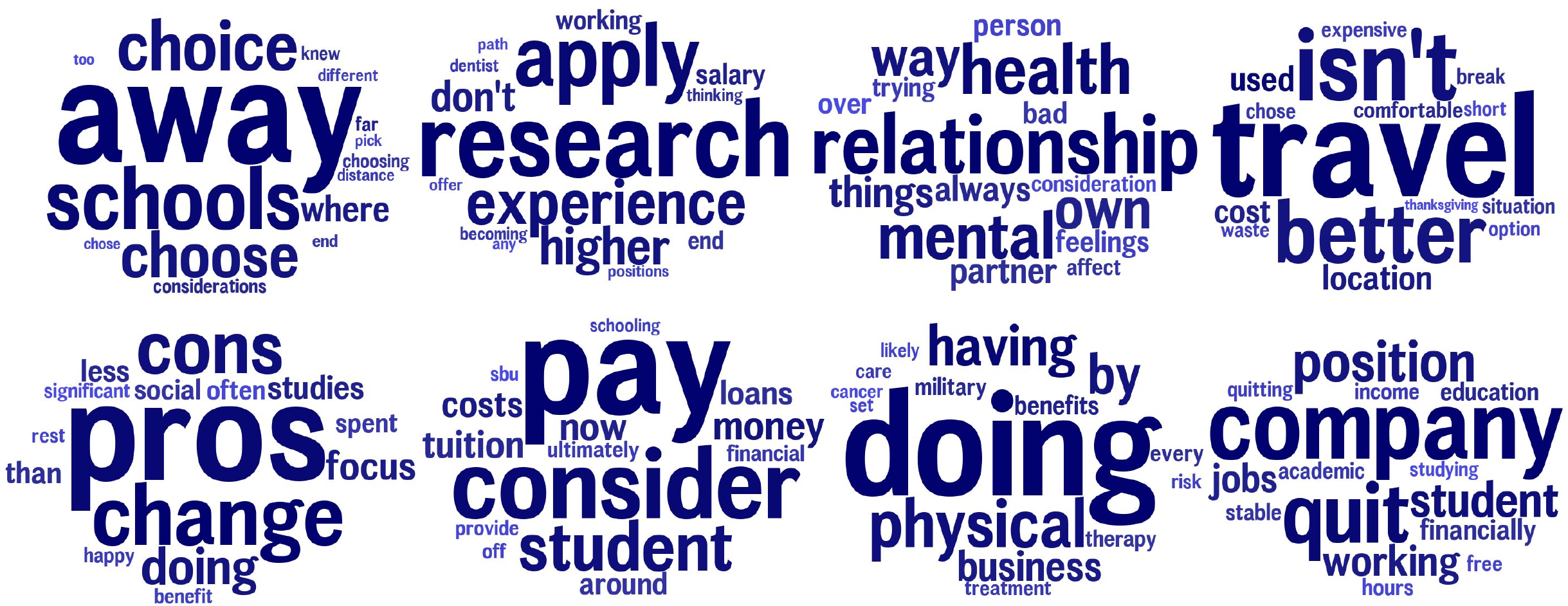}
  \caption{Randomly selected topics emerging from LDA on the participant writing describing a recent decision, depicting the types of content evoked. 
 }
  \label{fig:topics}
\end{figure}
\paragraph{Description of the \textit{Decisions} Dataset}
\label{sec:topics}
We employed topic modeling to describe the main themes of the \textit{Decisions} dataset while preserving privacy of our study participants. Figure~\ref{fig:topics} shows that respondents writing about their difficult decisions frequently mention topics related to college education, career goals, finances, mental health, friendships, family relationships, and vacation plans. These subjects are largely connected to common decision-making aspects of student life. The average length of essays is 186.28 words (min: 120, max: 508 words). The average Choice-Induced Shift (CIS) is 25.6  ($\sigma$: 38.4, min: -102.4, max: 140.8) 
-- this is consistent with the \citet{simon2004construction} paper that shows 
that more people tend to change their preferences towards the decision they make, i.e. CIS skews positive. Finally, out of the 502 participants, 417 (83\%) were influenced in the direction of the confounding attribute ($loc$) whereas 85 (17\%) remained uninfluenced.


 \section{Methods}

We explored the following theoretically relevant \textbf{Discourse Relations}:



\paragraph{1. Causal explanations} 
We extract individual reasoning behind decision-making behaviors using a causal explanation detection model trained on social media posts with a F1-macro of 0.85~\cite{son2018causal}. We infer the proportion of messages containing causal explanations provided by the individual.
\paragraph{2. Counterfactuals} These are statements of alternate reality; of what could have happened instead of actual events. We used the counterfactual relation recognition model based on a social media dataset with an F1-macro of 0.77 \cite{son-etal-2017-recognizing} to calculate the proportion of the messages from each individual that contains counterfactual statements.

\paragraph{3. Dissonance and Consonance} 
We extracted linguistic dissonance and consonance using a model trained on social media posts (AUC = 0.75) introduced in~\citet{varadarajan-etal-2023}, which captures signals of cognitive dissonance exhibited through language. We then calculated the average probability of dissonance for consecutive phrases predicted as dissonant or consonant.

\paragraph{4. Discourse Relation Embeddings} To capture \textit{other} discourse-level information, we use discourse relation embeddings that is extracted from pairs of consecutive discourse arguments~\cite{son-etal-2022-discourse}, aggregated by averaging at a message level.

 
Further, we explored common baseline models to capture the decision-making cognitive styles: a random baseline, zero- and four-shot prompting on both Llama3.1-8B-chat and Gemma-7B-Instruct
\footnote{The LLMs were prompted with the definitions of CIS and Inf variables. For the prompts, please check \S\ref{tab:prompts}.}, and finally, a predictive model from averaged embeddings of the text from L23 of RoBERTa-large.  

\paragraph{Predictive Models for Decision Making}
We model 2 outcomes together:  Choice-Induced Shift (CIS) and Influence (Inf). \textbf{CIS} and \textbf{Inf} variables capture the magnitude and direction of the tendency of a person to vacillate when exposed to conflict-inducing information. We combine them into a single variable \textbf{CIS\_Inf} for modeling four distinct cognitive styles for decision making: (a) Negative CIS,
Not Influenced (↓CIS↓Inf, 6\%), (b)
Negative CIS,
Influenced (↓CIS↑Inf, 17\%), (c) Positive CIS,
Not Influenced (↑CIS↓Inf, 11\%) and (d) Positive CIS,
Influenced (↑CIS↑Inf, 66\%).

We use a logistic regression model for 4-way classification with the features listed in Table~\ref{tab:CIS_modeling_results}, where we calculate stratified 5-fold cross-validation accuracies
using DLATK~\citep{schwartz2017dlatk}. 
\begin{table}[!ht]

\small
\begin{tabular}{l|r||l|lr}
\toprule
  

\multicolumn{1}{l}{\textbf{Baselines}} &
\textbf{AUC}
&\multicolumn{1}{l}{\textbf{Discourse feats}}  &
\multicolumn{1}{c}{\textbf{AUC}} 
& \textbf{k}\\
\midrule

Random & 0.50 &Causal 
& \textbf{0.81} 
 & 1\\
Llama3.1 (0-sh)  & 0.56 &Counterfactual 
& 0.80 
 & 1\\
Gemma (0-sh) &0.56 &Consonance  
& \textbf{0.81}
 & 1\\
Llama3.1 (4-sh)  &0.64&Dissonance  
& 0.80
 & 1\\
Gemma (4-sh) & \textbf{0.79}&\hspace{0mm}DiscRE (full) 
& 0.76 
 &845\\
RoBERTa-L23 & 0.69&\hspace{0mm}DiscRE (16-D) 
& 0.79
 &16\\

\bottomrule
\end{tabular}
\caption{Performance of various feature sets  over the CIS\_Inf outcome (\textbf{AUC}: mean Area Under the ROC Curve; \textbf{k}: number of input features). 
Linguistic measures from the participants' pre-experiment writing can predict CIS\_Inf with moderate-high, non-trivial accuracy.  
}
\label{tab:CIS_modeling_results}
\end{table}
\section{Results}
We explore results for our primary application of the \textit{experimental validation framework}: do discourse relation models, which capture explanatory styles and coherence in language of individuals, predict the cognitive style of a decision that an individual makes?
Table \ref{tab:CIS_modeling_results} shows that cognitive styles, represented by CIS\_Inf, have predictive correlates in language. CIS\_Inf captures two different variables (Fig 3) -- how much a person's preference shifts before and after the experiment and whether they were influenced in making the decision.
While discourse relation embeddings themselves seem to have low predictive power, specific relevant relations such as \textbf{Causal} and \textbf{Consonance} have high predictive power towards the cognitive styles of individuals pertaining to actual decision-making. 
With discourse relation features achieving an AUC of $\sim$0.8, language shows promise in capturing cognitive styles of individuals that are exhibited through their behavior. While few-shot prompting achieves comparable performance to discourse features, the latter's success is particularly noteworthy given their significantly lower parameter count compared to large language models. The effectiveness of these interpretable discourse features reinforces our finding that linguistic patterns reflect underlying cognitive styles.




\begin{table}[!ht]
\small
 \centering
 
 \begin{tabular}{p{0.4\linewidth}|p{0.09\linewidth}p{0.09\linewidth}p{0.09\linewidth}p{0.09\linewidth}}
 
 \toprule
\textbf{Theoretical Features} &\scriptsize{\textbf{$\downarrow$CIS$\downarrow$Inf}} & \scriptsize{\textbf{$\downarrow$CIS$\uparrow$Inf}} & \scriptsize
{\textbf{$\uparrow$CIS$\downarrow$Inf}}& \scriptsize{\textbf{$\uparrow$CIS$\uparrow$Inf}} \\

 \midrule
OCEAN
& \multicolumn{1}{l}{}          & \multicolumn{1}{l}{}          & \multicolumn{1}{l}{}          & \multicolumn{1}{l}{}          \\
 \hspace{5pt}Openness        & \cellcolor[HTML]{F6D2CF}-0.14 & \cellcolor[HTML]{B4E1CB}0.15  & \cellcolor[HTML]{F4C5C1}-0.18 & \cellcolor[HTML]{EAF7F1}0.03  \\
  \hspace{5pt}Conscientiousness & \cellcolor[HTML]{F6D2CF}-0.14 & \cellcolor[HTML]{F3C3BF}-0.18 & \cellcolor[HTML]{9ED8BB}0.20  & \cellcolor[HTML]{E1F3EA}0.05  \\
   \hspace{5pt}Extraversion      & \cellcolor[HTML]{FEFBFB}-0.03 & \cellcolor[HTML]{FFFFFF}-0.02 & \cellcolor[HTML]{E7F6EE}0.03  & \cellcolor[HTML]{F9FDFB}-0.01 \\
    \hspace{5pt}Agreeableness     & \cellcolor[HTML]{FCF3F3}-0.05 & \cellcolor[HTML]{B7E2CD}0.14  & \cellcolor[HTML]{E0F3EA}0.05  & \cellcolor[HTML]{FAE5E4}-0.09 \\
 \hspace{5pt}Emotional Stability         & \multicolumn{1}{c}{\cellcolor[HTML]{F6FCF9}{\color[HTML]{000000}0.00} } & \cellcolor[HTML]{F6FCF9}0.00  & \cellcolor[HTML]{FCF3F3}-0.05 & \cellcolor[HTML]{EFF9F4}0.02  \\

\midrule
Anxiety
& \cellcolor[HTML]{D5EEE2}0.07  & \cellcolor[HTML]{F9E1DE}-0.10 & \cellcolor[HTML]{88CFAC}0.24  & \cellcolor[HTML]{FCF0EF}-0.06 \\
Stress            & \cellcolor[HTML]{F4C9C5}-0.17 & \cellcolor[HTML]{DAF0E5}0.06  & \cellcolor[HTML]{D3EDE0}0.08  & \cellcolor[HTML]{FDF9F9}-0.04 \\
 Loneliness        & \cellcolor[HTML]{F3C1BD}-0.19 & \cellcolor[HTML]{F4C7C3}-0.17 & \cellcolor[HTML]{D4EEE1}0.08  & \cellcolor[HTML]{BFE5D3}0.12  \\
Empathic Concern     & \cellcolor[HTML]{AEDFC7}0.16  & \cellcolor[HTML]{C9E9D9}0.10  & \cellcolor[HTML]{F5CECB}-0.15 & \cellcolor[HTML]{FEFBFB}-0.03 \\
\midrule
Discourse  relations       & \multicolumn{1}{l}{}          & \multicolumn{1}{l}{}          & \multicolumn{1}{l}{}          & \multicolumn{1}{l}{}          \\
 \hspace{5pt}Causal            & \cellcolor[HTML]{73C79E}0.29  & \cellcolor[HTML]{F7D6D3}-0.13 & \cellcolor[HTML]{F5CECB}-0.15 & \cellcolor[HTML]{DBF1E6}0.06  \\
 \hspace{5pt}Counterfactual    & \cellcolor[HTML]{F8DCDA}-0.11 & \cellcolor[HTML]{F1F9F5}0.01  & \cellcolor[HTML]{BAE3CF}0.13  & \cellcolor[HTML]{FDF7F7}-0.04 \\
 \hspace{5pt}Consonance        & \cellcolor[HTML]{E3F4EC}0.04  & \cellcolor[HTML]{F6D0CC}-0.15 & \cellcolor[HTML]{F0B4AF}-0.22 & \cellcolor[HTML]{A4DBC0}0.18  \\
 \hspace{5pt}Dissonance        & \cellcolor[HTML]{A5DBC1}0.18  & \cellcolor[HTML]{F5CCC8}-0.16 & \cellcolor[HTML]{E7F5EE}0.04  & \cellcolor[HTML]{E8F6EF}0.03 \\

 \bottomrule
 \end{tabular}
 \caption{Cohen's \textit{d} for theoretical features against cognitive style outcomes of CIS\_Inf. 
 }

 \label{tab: correlations}
\end{table}

To explore language-specific patterns that relate to each type of cognitive style, we also extracted 
for theoretically-relevant lexical and discourse relation features in predicting each class of CIS\_Inf. Results are presented in Table~\ref{tab: correlations}, where we find that the four classes are highly differentiable along lexical-based measures for personality~\citep{park2015automatic}, anxiety~\citep{mangalik2024robust}, stress~\citep{guntuku2019understanding}, loneliness~\citep{guntuku2019studying} and empathic concern~\citep{giorgi2023human}. Discourse relations, especially the Causal relation has a Cohen's \textit{d} of 0.29 with the class ↓CIS↓Inf. 
We find that individuals who use more causal explanations and dissonant statements in their description of a recent past decision are less likely to change their minds about a decision due to external influence, and are less likely to change their preferences after making a decision, whereas, individuals who use less consonant statements in describing their decisions are more likely to switch their preferences after making a decision in the experiment. 

Interestingly, higher linguistic dissonance is associated with less change in preferences / tendency to be influenced, which may signal difficulty in resolving dissonance surrounding one's decision.
Higher change in preferences with low tendency to be influenced also seems to be signaled by linguistic anxiety, and each of the cognitive styles have a distinct signature across personality and well-being dimensions. 
This indicates that individual decision-making cognitive styles derived from simulated real-life experiments can be gleaned from personal discourse and the explanatory style of the person.


\paragraph{Recommendations:}

As an initial step in developing this evaluation framework, we recommend incorporating direct behavioral measurements into linguistic analyses, moving beyond traditional annotation-based methods. While annotations provide useful approximations of cognitive states, they rely on external judgments rather than direct psychological evidence. In contrast, experimental paradigms—widely used in psychology—allow researchers to systematically measure cognition and behavior under controlled conditions, offering a more reliable way to validate language-based models. To ensure ecological validity, language data should be collected before the experiment to prevent unintended influence on participants' responses.
To capture a fuller picture of cognitive processes, researchers should combine linguistic features with behavioral metrics such as response times (e.g., questionnaire completion speed), click-through rates, and dynamic shifts in participant responses. This multimodal approach provides stronger evidence for the relationship between language and cognition, allowing NLP models to be evaluated against real psychological processes rather than relying solely on subjective annotations. By integrating experimental methods, this framework strengthens the scientific grounding of language-based models and enhances their validity for applications in cognitive science, decision-making research, and human-computer interaction.

\section{Conclusion}

We demonstrated that experimentally-evoked cognitive styles can indeed be captured by language, offering a more solid ``ground truth'' compared to annotations of perceived behavior, which often fail to reflect a person's true state.
This framework emphasizes methodological rigor through controlled psychological experiments, enabling researchers to establish robust connections between language patterns and realistic estimates of cognitive states. 
Our framework's effectiveness is demonstrated with language-based features having strong predictive power for objective cognitive styles, especially discourse features successfully capturing experimentally measured cognitive styles. 
This approach not only enhances statistical validity but also has practical applications in the use of LLMs for mental health therapy, agent engagement systems, and cognitive science. 
By moving beyond the limitations of annotation-based or questionnaire-based labels, this paradigm represents a crucial step toward more rigorous evaluation in NLP, suggesting promising directions for future research in understanding the relationship between language and cognition.

\section*{Limitations}
While our experiment aims to capture cognitive dissonance through language in tandem with the replication of \citet{simon2004construction}, our study does not include direct questions in the writing prompts that explicitly prompt participants to discuss their decision-making process within the experiment itself. Despite the indirect writing prompt, we were able to capture promising cognitive style of individuals irrespective of the experimental outcome.  Further, the experiment offers a simulated job offer scenario, and the outcomes could be different in real-life. That said, our work is an initial step towards exploring associations of explicit linguistic structures and language modeling with observable psychological constructs, through the inclusion of psychological experiments in data collection. Therefore we chose a simpler abstraction of a real-world decision making problem as is usually done in the field of social psychology. However, this creates limitations in directly predicting participants' actual decision-making behaviors.

While discourse relations were originally intended to capture cognitive states through coherence and rhetorical structures, our predictive model-based method for inferring these relations offers only a small boost to the correlations when compared to lexical measures and contextual representations. This suggests that regular contextual embeddings might contain enough information to pick up cognitive styles and human behavior from language.

Our study population introduces several limitations that should be noted. The experiment uses undergraduate students at a public university which may limit the generalizability of the findings to other populations or age groups. While the study's focus on job decisions was particularly relevant to undergraduate students, who are often navigating a transitional phase focused career personal development, their decision-making processes may vary considerably from those of individuals in diverse life stages or professional environments. Furthermore, the linguistic outcomes were constrained by the small number of participants limited to the university. Therefore, the effect size was influenced by the restricted diversity in the population and the size of the participants.

\section*{Ethics Statement}

This study included an experiment with human subjects. The experiment followed closely to what that has been well replicated with no known risks in the past. The experiments were approved by ethical Institutional Review Board (IRB) who conducted a full review granting their approval. 

All participants provided informed consent prior to their participation. Participants were informed that they have the right to withdraw from the study at any time without any repercussions. Participants were also informed about how their data would be used and the measures taken to protect their privacy. Additionally participants confidentiality and privacy have been maintained throughout the research and analysis process. Any identifiable information collected during the study has been securely stored on a password-protected server, ensuring that only authorized personnel could access the information. All data were anonymized, any identifying details were removed or coded so that individuals could not be readily identified from the dataset. These steps ensured that the study upheld the highest ethical standards, prioritizing the privacy and well-being of all participants. The participants were paid USD 25 for completing the questionnaire after being recruited through the university. 

We run all of our experiments on an NVIDIA-RTX-A6000 with 50 GB of memory in an internal server, on open-sourced models. The LLMs were used for inferences rather than training for zero- and few-shot settings, with resource usage of about 15-20 hours on a single GPU. 

This work is part of a growing initiative to improve NLP for the human context. The models produced are not intended for any clinical or industrial application, and in particular not for targeted marketing or in use case where one's language is assessed for individual targeted information without individual awareness. 
The primary aim is to enhance the way cognitive processes are understood, ensuring that technology serves to augment psychological processes and measures. 

\section*{Acknowledgements}
This work was supported in part by a grant from the NIH-NIAAA (R01 AA028032) and a DARPA Young Faculty Award grant \#W911NF-20-1-0306 awarded to H. Andrew Schwartz at Stony Brook University.
The conclusions contained herein are those of the
authors and should not be interpreted as necessarily representing the official policies, either expressed or implied, of DARPA, NIH, any other
government organization, or the U.S. Government.

\bibliography{anthology,custom}

\begin{thebibliography}{46}
\expandafter\ifx\csname natexlab\endcsname\relax\def\natexlab#1{#1}\fi

\bibitem[{Aguilar et~al.(2022)Aguilar, Correia, de~Vries, and
  Tortora}]{aguilar2022cognitive}
Pilar Aguilar, Isabel Correia, Jan de~Vries, and Leda Tortora. 2022.
\newblock Cognitive dissonance induction as an “inoculator” against
  negative attitudes towards victims.
\newblock \emph{Social and Personality Psychology Compass}, 16(12):e12715.

\bibitem[{Alhamzeh et~al.(2022)Alhamzeh, Fonck, Versm{\'e}e, Egyed-Zsigmond,
  Kosch, and Brunie}]{alhamzeh2022s}
Alaa Alhamzeh, Romain Fonck, Erwan Versm{\'e}e, El{\"o}d Egyed-Zsigmond, Harald
  Kosch, and Lionel Brunie. 2022.
\newblock It’s time to reason: Annotating argumentation structures in
  financial earnings calls: The finarg dataset.
\newblock In \emph{Proceedings of the Fourth Workshop on Financial Technology
  and Natural Language Processing (FinNLP)}, pages 163--169.

\bibitem[{Brook~O’Donnell and Falk(2015)}]{brook2015linking}
Matthew Brook~O’Donnell and Emily~B Falk. 2015.
\newblock Linking neuroimaging with functional linguistic analysis to
  understand processes of successful communication.
\newblock \emph{Communication Methods and Measures}, 9(1-2):55--77.

\bibitem[{Buchanan et~al.(2013)Buchanan, Seligman, and
  Seligman}]{buchanan2013explanatory}
Gregory~McClell Buchanan, Martin~EP Seligman, and Martin Seligman. 2013.
\newblock \emph{Explanatory style}.
\newblock Routledge.

\bibitem[{Buechel et~al.(2018)Buechel, Buffone, Slaff, Ungar, and
  Sedoc}]{buechel-etal-2018-modeling}
Sven Buechel, Anneke Buffone, Barry Slaff, Lyle Ungar, and Jo{\~a}o Sedoc.
  2018.
\newblock \href {https://doi.org/10.18653/v1/D18-1507} {Modeling empathy and
  distress in reaction to news stories}.
\newblock In \emph{Proceedings of the 2018 Conference on Empirical Methods in
  Natural Language Processing}, pages 4758--4765, Brussels, Belgium.
  Association for Computational Linguistics.

\bibitem[{Campitelli and Gobet(2010)}]{campitelli2010herbert}
Guillermo Campitelli and Fernand Gobet. 2010.
\newblock Herbert simon's decision-making approach: Investigation of cognitive
  processes in experts.
\newblock \emph{Review of general psychology}, 14(4):354--364.

\bibitem[{Choi et~al.(2023)Choi, Pei, Kumar, Shu, and
  Jurgens}]{choi-etal-2023-llms}
Minje Choi, Jiaxin Pei, Sagar Kumar, Chang Shu, and David Jurgens. 2023.
\newblock \href {https://doi.org/10.18653/v1/2023.emnlp-main.699} {Do {LLM}s
  understand social knowledge? evaluating the sociability of large language
  models with {S}oc{KET} benchmark}.
\newblock In \emph{Proceedings of the 2023 Conference on Empirical Methods in
  Natural Language Processing}, pages 11370--11403, Singapore. Association for
  Computational Linguistics.

\bibitem[{Dey et~al.(2024)Dey, Ganesan, Lal, Shah, Sinha, Matero, Giorgi,
  Kulkarni, and Schwartz}]{dey2024socialite}
Gourab Dey, Adithya~V Ganesan, Yash~Kumar Lal, Manal Shah, Shreyashee Sinha,
  Matthew Matero, Salvatore Giorgi, Vivek Kulkarni, and H~Andrew Schwartz.
  2024.
\newblock Socialite-llama: An instruction-tuned model for social scientific
  tasks.
\newblock \emph{arXiv preprint arXiv:2402.01980}.

\bibitem[{Giorgi et~al.(2023)Giorgi, Havaldar, Ahmed, Akhtar, Vaidya, Pan,
  Ungar, Schwartz, and Sedoc}]{giorgi2023human}
Salvatore Giorgi, Shreya Havaldar, Farhan Ahmed, Zuhaib Akhtar, Shalaka Vaidya,
  Gary Pan, Lyle~H Ungar, H~Andrew Schwartz, and Joao Sedoc. 2023.
\newblock Human-centered metrics for dialog system evaluation.
\newblock \emph{arXiv preprint arXiv:2305.14757}.

\bibitem[{Guntuku et~al.(2019{\natexlab{a}})Guntuku, Buffone, Jaidka,
  Eichstaedt, and Ungar}]{guntuku2019understanding}
Sharath~Chandra Guntuku, Anneke Buffone, Kokil Jaidka, Johannes~C Eichstaedt,
  and Lyle~H Ungar. 2019{\natexlab{a}}.
\newblock Understanding and measuring psychological stress using social media.
\newblock In \emph{Proceedings of the international AAAI conference on web and
  social media}, volume~13, pages 214--225.

\bibitem[{Guntuku et~al.(2019{\natexlab{b}})Guntuku, Schneider, Pelullo, Young,
  Wong, Ungar, Polsky, Volpp, and Merchant}]{guntuku2019studying}
Sharath~Chandra Guntuku, Rachelle Schneider, Arthur Pelullo, Jami Young, Vivien
  Wong, Lyle Ungar, Daniel Polsky, Kevin~G Volpp, and Raina Merchant.
  2019{\natexlab{b}}.
\newblock Studying expressions of loneliness in individuals using twitter: an
  observational study.
\newblock \emph{BMJ open}, 9(11):e030355.

\bibitem[{Harmon-Jones and Harmon-Jones(2007)}]{harmon2007cognitive}
Eddie Harmon-Jones and Cindy Harmon-Jones. 2007.
\newblock Cognitive dissonance theory after 50 years of development.
\newblock \emph{Zeitschrift f{\"u}r Sozialpsychologie}, 38(1):7--16.

\bibitem[{Hayati et~al.(2021)Hayati, Kang, and Ungar}]{hayati-etal-2021-bert}
Shirley~Anugrah Hayati, Dongyeop Kang, and Lyle Ungar. 2021.
\newblock \href {https://doi.org/10.18653/v1/2021.emnlp-main.510} {Does {BERT}
  learn as humans perceive? understanding linguistic styles through lexica}.
\newblock In \emph{Proceedings of the 2021 Conference on Empirical Methods in
  Natural Language Processing}, pages 6323--6331, Online and Punta Cana,
  Dominican Republic. Association for Computational Linguistics.

\bibitem[{Juhng et~al.(2023)Juhng, Matero, Varadarajan, Eichstaedt, Ganesan,
  and Schwartz}]{juhng2023discourse}
Swanie Juhng, Matthew Matero, Vasudha Varadarajan, Johannes Eichstaedt,
  Adithya~V Ganesan, and H~Andrew Schwartz. 2023.
\newblock Discourse-level representations can improve prediction of degree of
  anxiety.
\newblock In \emph{Proceedings of the 61st Annual Meeting of the Association
  for Computational Linguistics (Volume 2: Short Papers)}, pages 1500--1511.

\bibitem[{Knaebel and Stede(2023)}]{knaebel2023discourse}
Ren{\'e} Knaebel and Manfred Stede. 2023.
\newblock Discourse sense flows: Modelling the rhetorical style of documents
  across various domains.
\newblock In \emph{Findings of the Association for Computational Linguistics:
  EMNLP 2023}, pages 14462--14482.

\bibitem[{Lahnala et~al.(2022)Lahnala, Welch, Jurgens, and
  Flek}]{lahnala-etal-2022-critical}
Allison Lahnala, Charles Welch, David Jurgens, and Lucie Flek. 2022.
\newblock \href {https://doi.org/10.18653/v1/2022.findings-emnlp.157} {A
  critical reflection and forward perspective on empathy and natural language
  processing}.
\newblock In \emph{Findings of the Association for Computational Linguistics:
  EMNLP 2022}, pages 2139--2158, Abu Dhabi, United Arab Emirates. Association
  for Computational Linguistics.

\bibitem[{Lahnala et~al.(2024)Lahnala, Neuendorf, Thomin, Welch, Stibane, and
  Flek}]{lahnala-etal-2024-appraisal-framework}
Allison~Claire Lahnala, B{\'e}la Neuendorf, Alexander Thomin, Charles Welch,
  Tina Stibane, and Lucie Flek. 2024.
\newblock \href {https://aclanthology.org/2024.lrec-main.124} {Appraisal
  framework for clinical empathy: A novel application to breaking bad news
  conversations}.
\newblock In \emph{Proceedings of the 2024 Joint International Conference on
  Computational Linguistics, Language Resources and Evaluation (LREC-COLING
  2024)}, pages 1393--1407, Torino, Italia. ELRA and ICCL.

\bibitem[{Liang et~al.(2020)Liang, Zou, and Yu}]{liang2020beyond}
Weixin Liang, James Zou, and Zhou Yu. 2020.
\newblock Beyond user self-reported likert scale ratings: A comparison model
  for automatic dialog evaluation.
\newblock \emph{arXiv preprint arXiv:2005.10716}.

\bibitem[{Mangalik et~al.(2024)Mangalik, Eichstaedt, Giorgi, Mun, Ahmed, Gill,
  V.~Ganesan, Subrahmanya, Soni, Clouston et~al.}]{mangalik2024robust}
Siddharth Mangalik, Johannes~C Eichstaedt, Salvatore Giorgi, Jihu Mun, Farhan
  Ahmed, Gilvir Gill, Adithya V.~Ganesan, Shashanka Subrahmanya, Nikita Soni,
  Sean~AP Clouston, et~al. 2024.
\newblock Robust language-based mental health assessments in time and space
  through social media.
\newblock \emph{NPJ Digital Medicine}, 7(1):109.

\bibitem[{McGrath(2017)}]{mcgrath2017dealing}
April McGrath. 2017.
\newblock Dealing with dissonance: A review of cognitive dissonance reduction.
\newblock \emph{Social and Personality Psychology Compass}, 11(12):e12362.

\bibitem[{Meaney et~al.(2021)Meaney, Wilson, Chiruzzo, Lopez, and
  Magdy}]{meaney2021semeval}
JA~Meaney, Steven Wilson, Luis Chiruzzo, Adam Lopez, and Walid Magdy. 2021.
\newblock Semeval 2021 task 7: Hahackathon, detecting and rating humor and
  offense.
\newblock In \emph{Proceedings of the 15th International Workshop on Semantic
  Evaluation (SemEval-2021)}, pages 105--119.

\bibitem[{Mohammad et~al.(2018)Mohammad, Bravo-Marquez, Salameh, and
  Kiritchenko}]{mohammad2018semeval}
Saif Mohammad, Felipe Bravo-Marquez, Mohammad Salameh, and Svetlana
  Kiritchenko. 2018.
\newblock Semeval-2018 task 1: Affect in tweets.
\newblock In \emph{Proceedings of the 12th international workshop on semantic
  evaluation}, pages 1--17.

\bibitem[{Nisbett and Wilson(1977)}]{nisbett1977halo}
Richard~E Nisbett and Timothy~D Wilson. 1977.
\newblock The halo effect: Evidence for unconscious alteration of judgments.
\newblock \emph{Journal of personality and social psychology}, 35(4):250.

\bibitem[{Park et~al.(2015)Park, Schwartz, Eichstaedt, Kern, Kosinski,
  Stillwell, Ungar, and Seligman}]{park2015automatic}
Gregory Park, H~Andrew Schwartz, Johannes~C Eichstaedt, Margaret~L Kern, Michal
  Kosinski, David~J Stillwell, Lyle~H Ungar, and Martin~EP Seligman. 2015.
\newblock Automatic personality assessment through social media language.
\newblock \emph{Journal of personality and social psychology}, 108(6):934.

\bibitem[{Paulhus et~al.(2007)Paulhus, Vazire et~al.}]{paulhus2007self}
Delroy~L Paulhus, Simine Vazire, et~al. 2007.
\newblock The self-report method.
\newblock \emph{Handbook of research methods in personality psychology},
  1(2007):224--239.

\bibitem[{Rosenthal et~al.(2019)Rosenthal, Farra, and
  Nakov}]{rosenthal2019semeval}
Sara Rosenthal, Noura Farra, and Preslav Nakov. 2019.
\newblock Semeval-2017 task 4: Sentiment analysis in twitter.
\newblock \emph{arXiv preprint arXiv:1912.00741}.

\bibitem[{Sandri et~al.(2023)Sandri, Leonardelli, Tonelli, and
  Jezek}]{sandri-etal-2023-dont}
Marta Sandri, Elisa Leonardelli, Sara Tonelli, and Elisabetta Jezek. 2023.
\newblock \href {https://doi.org/10.18653/v1/2023.eacl-main.178} {Why don{'}t
  you do it right? analysing annotators{'} disagreement in subjective tasks}.
\newblock In \emph{Proceedings of the 17th Conference of the European Chapter
  of the Association for Computational Linguistics}, Dubrovnik, Croatia.
  Association for Computational Linguistics.

\bibitem[{Sap et~al.(2021)Sap, Swayamdipta, Vianna, Zhou, Choi, and
  Smith}]{sap2021annotators}
Maarten Sap, Swabha Swayamdipta, Laura Vianna, Xuhui Zhou, Yejin Choi, and
  Noah~A Smith. 2021.
\newblock Annotators with attitudes: How annotator beliefs and identities bias
  toxic language detection.
\newblock \emph{arXiv preprint arXiv:2111.07997}.

\bibitem[{Saxbe et~al.(2013)Saxbe, Yang, Borofsky, and
  Immordino-Yang}]{saxbe2013embodiment}
Darby~E Saxbe, Xiao-Fei Yang, Larissa~A Borofsky, and Mary~Helen
  Immordino-Yang. 2013.
\newblock The embodiment of emotion: language use during the feeling of social
  emotions predicts cortical somatosensory activity.
\newblock \emph{Social cognitive and affective neuroscience}, 8(7):806--812.

\bibitem[{Schwartz et~al.(2017)Schwartz, Giorgi, Sap, Crutchley, Ungar, and
  Eichstaedt}]{schwartz2017dlatk}
H~Andrew Schwartz, Salvatore Giorgi, Maarten Sap, Patrick Crutchley, Lyle
  Ungar, and Johannes Eichstaedt. 2017.
\newblock Dlatk: Differential language analysis toolkit.
\newblock In \emph{Proceedings of the 2017 conference on empirical methods in
  natural language processing: System demonstrations}, pages 55--60.

\bibitem[{Sharma et~al.(2020)Sharma, Miner, Atkins, and
  Althoff}]{sharma2020computational}
Ashish Sharma, Adam~S Miner, David~C Atkins, and Tim Althoff. 2020.
\newblock A computational approach to understanding empathy expressed in
  text-based mental health support.
\newblock \emph{arXiv preprint arXiv:2009.08441}.

\bibitem[{Sharma et~al.(2023)Sharma, Rushton, Lin, Wadden, Lucas, Miner,
  Nguyen, and Althoff}]{sharma-etal-2023-cognitive}
Ashish Sharma, Kevin Rushton, Inna Lin, David Wadden, Khendra Lucas, Adam
  Miner, Theresa Nguyen, and Tim Althoff. 2023.
\newblock \href {https://doi.org/10.18653/v1/2023.acl-long.555} {Cognitive
  reframing of negative thoughts through human-language model interaction}.
\newblock In \emph{Proceedings of the 61st Annual Meeting of the Association
  for Computational Linguistics (Volume 1: Long Papers)}, pages 9977--10000,
  Toronto, Canada. Association for Computational Linguistics.

\bibitem[{Simon et~al.(2004)Simon, Krawczyk, and
  Holyoak}]{simon2004construction}
Dan Simon, Daniel~C Krawczyk, and Keith~J Holyoak. 2004.
\newblock Construction of preferences by constraint satisfaction.
\newblock \emph{Psychological Science}, 15(5):331--336.

\bibitem[{Son et~al.(2018{\natexlab{a}})Son, Bayas, and
  Schwartz}]{son-etal-2018-causal}
Youngseo Son, Nipun Bayas, and H.~Andrew Schwartz. 2018{\natexlab{a}}.
\newblock \href {https://doi.org/10.18653/v1/D18-1372} {Causal explanation
  analysis on social media}.
\newblock In \emph{Proceedings of the 2018 Conference on Empirical Methods in
  Natural Language Processing}, pages 3350--3359, Brussels, Belgium.
  Association for Computational Linguistics.

\bibitem[{Son et~al.(2018{\natexlab{b}})Son, Bayas, and
  Schwartz}]{son2018causal}
Youngseo Son, Nipun Bayas, and H.~Andrew Schwartz. 2018{\natexlab{b}}.
\newblock Causal explanation analysis on social media.
\newblock In \emph{Proceedings of the 2018 Conference on Empirical Methods in
  Natural Language Processing}. Association for Computational Linguistics.

\bibitem[{Son et~al.(2017)Son, Buffone, Raso, Larche, Janocko, Zembroski,
  Schwartz, and Ungar}]{son-etal-2017-recognizing}
Youngseo Son, Anneke Buffone, Joe Raso, Allegra Larche, Anthony Janocko, Kevin
  Zembroski, H~Andrew Schwartz, and Lyle Ungar. 2017.
\newblock \href {https://doi.org/10.18653/v1/P17-2103} {Recognizing
  counterfactual thinking in social media texts}.
\newblock In \emph{Proceedings of the 55th Annual Meeting of the Association
  for Computational Linguistics (Volume 2: Short Papers)}, pages 654--658,
  Vancouver, Canada. Association for Computational Linguistics.

\bibitem[{Son et~al.(2022)Son, Varadarajan, and
  Schwartz}]{son-etal-2022-discourse}
Youngseo Son, Vasudha Varadarajan, and H.~Andrew Schwartz. 2022.
\newblock \href {https://aclanthology.org/2022.umios-1.5} {Discourse relation
  embeddings: Representing the relations between discourse segments in social
  media}.
\newblock In \emph{Proceedings of the Workshop on Unimodal and Multimodal
  Induction of Linguistic Structures (UM-IoS)}, pages 45--55, Abu Dhabi, United
  Arab Emirates (Hybrid). Association for Computational Linguistics.

\bibitem[{Taboada and Mann(2006)}]{taboada2006rhetorical}
Maite Taboada and William~C Mann. 2006.
\newblock Rhetorical structure theory: Looking back and moving ahead.
\newblock \emph{Discourse studies}, 8(3):423--459.

\bibitem[{Valizadeh et~al.(2021)Valizadeh, Ranjbar-Noiey, Caragea, and
  Parde}]{valizadeh-etal-2021-identifying}
Mina Valizadeh, Pardis Ranjbar-Noiey, Cornelia Caragea, and Natalie Parde.
  2021.
\newblock \href {https://doi.org/10.18653/v1/2021.naacl-main.347} {Identifying
  medical self-disclosure in online communities}.
\newblock In \emph{Proceedings of the 2021 Conference of the North American
  Chapter of the Association for Computational Linguistics: Human Language
  Technologies}, pages 4398--4408, Online. Association for Computational
  Linguistics.

\bibitem[{Van~Dijk(1990)}]{van1990social}
Teun~A Van~Dijk. 1990.
\newblock Social cognition and discourse.
\newblock \emph{Handbook of language and social psychology}, 163:183.

\bibitem[{Van~Dijk(2014)}]{van2014discourse}
Teun~A Van~Dijk. 2014.
\newblock Discourse, cognition, society.
\newblock \emph{The discourse studies reader: Main currents in theory and
  analysis}, page 388.

\bibitem[{Varadarajan et~al.(2023)Varadarajan, Juhng, Mahwish, Liu, Luby,
  Luhmann, and Schwartz}]{varadarajan-etal-2023}
Vasudha Varadarajan, Swanie Juhng, Syeda Mahwish, Xiaoran Liu, Jonah Luby,
  Christian~C. Luhmann, and H.~Andrew Schwartz. 2023.
\newblock Transfer and active learning for dissonance detection: Addressing the
  rare-class challenge.
\newblock In \emph{Proceedings of The 61st Annual Meeting of the Association
  for Computational Linguistics}. Association for Computational Linguistics.

\bibitem[{Varadarajan et~al.(2024)Varadarajan, Lahnala, V~Ganesan, Dey,
  Mangalik, Bucur, Soni, Rao, Lanning, Vallejo, Flek, Schwartz, Welch, and
  Boyd}]{varadarajan-etal-2024-archetypes}
Vasudha Varadarajan, Allison Lahnala, Adithya V~Ganesan, Gourab Dey, Siddharth
  Mangalik, Ana-Maria Bucur, Nikita Soni, Rajath Rao, Kevin Lanning, Isabella
  Vallejo, Lucie Flek, H.~Andrew Schwartz, Charles Welch, and Ryan Boyd. 2024.
\newblock \href {https://aclanthology.org/2024.clpsych-1.28/} {Archetypes and
  entropy: Theory-driven extraction of evidence for suicide risk}.
\newblock In \emph{Proceedings of the 9th Workshop on Computational Linguistics
  and Clinical Psychology (CLPsych 2024)}, pages 278--291, St. Julians, Malta.
  Association for Computational Linguistics.

\bibitem[{Varadarajan et~al.(2022)Varadarajan, Soni, Wang, Luhmann, Schwartz,
  and Inoue}]{varadarajan2022disso}
Vasudha Varadarajan, Nikita Soni, Weixi Wang, Christian Luhmann, H.~Andrew
  Schwartz, and Naoya Inoue. 2022.
\newblock \href {https://aclanthology.org/2022.nlpcss-1.16} {Detecting
  dissonant stance in social media: The role of topic exposure}.
\newblock In \emph{Proceedings of the Fifth Workshop on Natural Language
  Processing and Computational Social Science (NLP+CSS)}. Association for
  Computational Linguistics.

\bibitem[{Yang et~al.(2021)Yang, Hooshmand, and Hirschberg}]{yang2021choral}
Zixiaofan Yang, Shayan Hooshmand, and Julia Hirschberg. 2021.
\newblock Choral: Collecting humor reaction labels from millions of social
  media users.
\newblock In \emph{Proceedings of the 2021 Conference on Empirical Methods in
  Natural Language Processing}, pages 4429--4435.

\bibitem[{Zirikly et~al.(2019)Zirikly, Resnik, Uzuner, and
  Hollingshead}]{zirikly-etal-2019-clpsych}
Ayah Zirikly, Philip Resnik, {\"O}zlem Uzuner, and Kristy Hollingshead. 2019.
\newblock \href {https://doi.org/10.18653/v1/W19-3003} {{CLP}sych 2019 shared
  task: Predicting the degree of suicide risk in {R}eddit posts}.
\newblock In \emph{Proceedings of the Sixth Workshop on Computational
  Linguistics and Clinical Psychology}, pages 24--33, Minneapolis, Minnesota.
  Association for Computational Linguistics.

\end{thebibliography}
\bibliographystyle{acl_natbib}

\clearpage
\appendix
\section*{Appendix}
\counterwithin{figure}{section}
\counterwithin{table}{section}
\label{sec:appendix}

  
\begin{figure}
 \centering
 \includegraphics[width=\columnwidth]{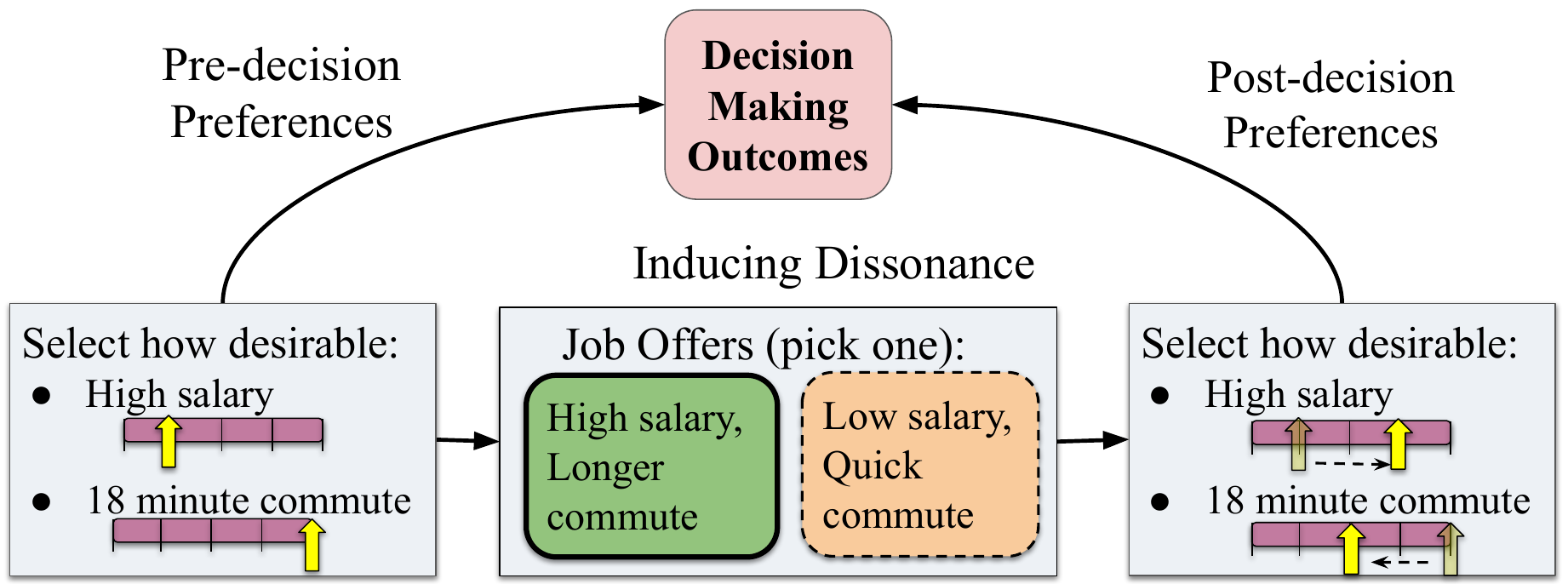}

 \caption{
After participants wrote about recent decisions that they had made (Step 1 in Figure 1), 
they completed a decision-making experiment wherein they encountered a simulated a job offer setting 
 (See \S\ref{subsec:job_offer}). If the participant picks the job with higher salary and longer commute (marked in green), their preferences are expected to change in the direction of preferring high salary more, and less in the direction of preferring short commute times. 
 }
 \label{fig:simons_exp}
\end{figure}

\section{Job Offer Questions}
A schematic diagram to demonstrate how the preference change is measured is shown in Figure~\ref{fig:simons_exp}. The detailed questionnaire administered to the participants is shown in Table \ref{tab:job_offer_qus}.
\label{sec:joboffer_desc}
\begin{table*}[]
\small
\begin{tabular}{p{0.8\linewidth}p{0.1\linewidth}}
\toprule
\textbf{Questions} & \textbf{Response type} \\
\midrule
\textbf{Writing about a recent difficult decision }\\
\hspace{1mm} 1. Please describe a recent important and difficult decision that you have made (20-150 words) & text \\
\hspace{1mm} 2. What were the considerations that you thought about while making the decision? When answering, please consider all of the circumstances and details that went into the difficult decision (100-300 words) & text \\
\midrule
\textbf{Background:}\\Imagine that you have just graduated from college and have decided to look for a job. You have had interviews with a few companies, and are hoping to receive some job offers. In this experiment you will be asked to state how you feel about an assortment of aspects that might be included in job offers. 
Specifically, you will be asked to state how desirable or undesirable you find each aspect. 
There are no right or wrong answers to these questions. Please state how you personally feel about these aspects as if you were evaluating them in the context of making a real decision about your future career. You are not expected to have any special knowledge.
You might find that the information given to you is less complete than you would like to have; nonetheless, respond as best as you can given the available information.
The issues are unrelated, so simply consider each one independently.
\\
 \midrule
\hspace{1mm} 1. A company maintains a national training center in Jackstown, Tennessee. Every employee must spend 3 weeks of training at that center every year. Most employees describe the training as boring and the life in Jackstown as gloomy. - Please select how desirable participating in the training sessions at Jackstown is to you. & -5 to 5 \\
\hspace{1mm} 2. The commute to work will take you about 18 minutes each way. - Please select how desirable the 18 minute commute is to you. & -5 to 5 \\
\hspace{1mm} 3. The average annual salary for the position you are considering is \$60,000. The salary you are being offered is \$61,200. - Please select how desirable it is to you to receive \$1200 above the average salary. & -5 to 5 \\
\hspace{1mm} 4. You will be given a cubicle, which is located in a pretty noisy area. - Please select how desirable it is to work in a cubicle. & -5 to 5 \\
\hspace{1mm} 5. Given your credentials, you should be considered for promotion within a year or two. Being promoted will mean that you will have more independence, but it also means that you will have many more responsibilities. Some veterans maintain that in this type of profession, it is best to gain more experience before being promoted. - Please select how desirable a promotion is to you. & -5 to 5 \\
\hspace{1mm}6. All companies give their employees at least two weeks of vacation a year.  Some companies give additional vacation benefits.  A company offers you only the minimum two-week vacation. - Please select how desirable it is to receive only the minimum two-week vacation. & -5 to 5 \\
\hspace{1mm}7. The commute to work will take you about 40 minutes each way. - Please select how desirable the 40 minute commute is to you. & -5 to 5 \\
\hspace{1mm} 8. You are offered an office to yourself.  The office is pretty small, though adequate. - Please select how desirable the private office is to you. & -5 to 5 \\
\hspace{1mm}9. The average annual salary for the position you are considering is \$60,000. A company offers you \$59,100. - Please select how desirable it is to you to receive \$900 below the average salary. & -5 to 5 \\
\hspace{1mm}10. In addition to the standard two-week annual vacation, a company takes its employees and their families to a week-long retreat in San Diego.  The retreat consists of work-related lectures and workshops, but it is usually quite a lot of fun. - Please select how desirable the retreat in San Diego is to you. & -5 to 5 \\
\hspace{1mm}11. A company has a policy of encouraging personnel mobility among its numerous branches located throughout the country and across Europe.  Every employee is entitled to spend up to 3 months every 2 years working at any one of the company's branches. - Please select how desirable this mobility is to you. & -5 to 5 \\
\midrule
1. Please state the relative weight you would assign each of the aspects in the overall context of choosing a job (using the slider).  You are encouraged to use the full range of the scale: - 1. The office & 1 to 8 \\
2. Please state the relative weight you would assign each of the aspects in the overall context of choosing a job (using the slider).  You are encouraged to use the full range of the scale: - 2. The commute & 1 to 8 \\
3. Please state the relative weight you would assign each of the aspects in the overall context of choosing a job (using the slider).  You are encouraged to use the full range of the scale: - 3. The salary & 1 to 8 \\
4. Please state the relative weight you would assign each of the aspects in the overall context of choosing a job (using the slider).  You are encouraged to use the full range of the scale: - 4. The vacation package & 1 to 8 \\
\bottomrule
\end{tabular}

\end{table*}
\begin{table*}
\small
\begin{tabular}{p{0.98\linewidth}}

\midrule 
\textbf{[DISTRACTION] Synonyms task: Match the synonyms for 20 moderately difficult English words}   \\
\midrule
\textbf{Background:}\\
In this experiment you will be asked to play the role of a person who has just graduated from college. You are currently looking for a job in the field of marketing.You have just received interesting job offers from two large department store chains, Splendor and Bonnie’s Best. The two companies are similar in terms of their size, reputation and stability, and your prospects for promotion seem the same with both companies. You have already spent a couple of days at each of their offices, and have been interviewed by the key personnel. You found both companies to be stimulating and pleasant. After receiving more information about the two job offers, you will be asked to decide which one to accept.\\
\bottomrule
\end{tabular}

\textbf{Participants randomly get one of the two configurations (one with Splendor in a positive $loc$ condition and the other with Bonnie's Best in a positive $loc$ condition):}\\
\begin{tabular}{p{0.48\linewidth}||p{0.48\linewidth}}
\toprule

\textbf{Option A: Splendor (positive $loc$ condition)}& \textbf{Option A: Bonnie’s Best (positive $loc$ condition)}\\
Splendor is located in a fun part of town, next door to a new mall. There are many food joints, clothing stores, and cinemas close by. Most of the employees there go out to lunch in groups and eat at different places every day. They also do some convenient shopping on their way home from work. The average annual salary of a person at your position is \$60,000. The salary you are being offered by Splendor is \$59,100. At Splendor, you are offered an office to yourself. The office is pretty small, though adequate.The commute to the offices of Splendor takes about 18 minutes each way. Splendor offers its employees two weeks of vacation a year.&   Bonnie's Best is located in a fun part of town, next door to a new mall. There are many food joints, clothing stores, and cinemas close by. Most of the employees there go out to lunch in groups and eat at different places every day. They also do some convenient shopping on their way home from work. The average annual salary of a person at your position is \$60,000. The salary you are being offered by Bonnie’s Best is \$61,200. At Bonnie’s Best, you will be given a cubicle, which is located in a pretty noisy area. The commute to the offices of Bonnie’s Best takes about 40 minutes each way. In addition to the standard two-week annual vacation, every summer Bonnie’s Best takes its employees and their families to a retreat in San Diego. The retreat consists of work-related lectures and workshops, but it is usually quite a lot of fun.
\vspace{5mm}\\ \textbf{Option B: Bonnie’s Best} & \textbf{Option B: Splendor}\\Bonnie’s Best is located in a dull, sparsely populated industrial area. There is only one mediocre cafeteria nearby. Most employees bring their own sandwiches and eat on their own, or spend much of their lunch break driving to eateries that are a fair distance away. The average annual salary of a person at your position is \$60,000. The salary you are being offered by Bonnie’s Best is \$61,200. At Bonnie’s Best, you will be given a cubicle, which is located in a pretty noisy area. The commute to the offices of Bonnie’s Best takes about 40 minutes each way. In addition to the standard two-week annual vacation, every summer Bonnie’s Best takes its employees and their families to a retreat in San Diego. The retreat consists of work-related lectures and workshops, but it is usually quite a lot of fun.&Splendor is located in a dull, sparsely populated industrial area. There is only one mediocre cafeteria nearby. Most employees bring their own sandwiches and eat on their own, or spend much of their lunch break driving to eateries that are a fair distance away. The average annual salary of a person at your position is \$60,000. The salary you are being offered by Splendor is \$59,100. At Splendor, you are offered an office to yourself. The office is pretty small, though adequate. The commute to the offices of Splendor takes about 18 minutes each way. Splendor offers its employees two weeks of vacation a year. \\
\bottomrule
\end{tabular}
\end{table*}
\begin{table*}
\small
\begin{tabular}{p{0.8\linewidth}p{0.1\linewidth}}
\toprule
\textbf{Questions} & \textbf{Response type} \\
\midrule
At this point you have all the available information, and you are now asked to make your decision. Take your time and feel free to look back at the information provided. Please consider all pros and cons of both job offers carefully. Try to make this decision as if you were really in the described situation, and were facing a choice that will strongly influence your future career. When you have made your decision, please choose one of the two options. I accept the job offer of: & Bonnie's Best /  Splendor \\
\midrule
You will now be requested to state your preferences towards the aspects of the job offers made by Splendor and Bonnie’s Best. Specifically, you are requested to state how desirable or undesirable you find each of these aspects. There are no right or wrong answers to these questions. Please state your subjective preferences.
You are requested to answer the following questions using the provided scales. You are encouraged to use the full range of the scale:
\\\\
\hspace{1mm} 1.The commute to the offices of Splendor takes about 18 minutes each way. - Please select how desirable the 18 minute commute is to you. & -5 to 5 \\
\hspace{1mm} 2. Splendor does not offer any vacation benefits above the minimum two-week vacation a year. - Please select how desirable it is to receive only the minimum two-week vacation. & -5 to 5 \\
\hspace{1mm} 3.The salary you are being offered by Bonnie’s Best is \$1,200 above the average salary in the field. - Please select how desirable it is to you to receive \$1200 above the average salary. & -5 to 5 \\
\hspace{1mm} 4. At Splendor, you are offered an office to yourself.  The office is pretty small, though adequate. - Please select how desirable the private office is to you. & -5 to 5 \\
\hspace{1mm} 5. At Bonnie’s Best, you will be given a cubicle, which is located in a pretty noisy area. - Please select how desirable it is to work in a cubicle. & -5 to 5 \\
\hspace{1mm} 6. In addition to the standard two-week annual vacation, every summer Bonnie’s Best takes its employees and their families to a retreat in San Diego.  The retreat consists of work-related lectures and workshops, but it is usually quite a lot of fun. - Please select how desirable the San Diego retreat is to you. & -5 to 5 \\
\hspace{1mm} 7. The commute to the offices of Bonnie’s Best takes about 40 minutes each way. - Please select how desirable the 40 minute commute is to you. & -5 to 5 \\
\hspace{1mm} 8. The salary you are being offered by Splendor is \$900 below the average salary in the field. - Please select how desirable it is to you to receive \$900 below the average salary. & -5 to 5 \\
\bottomrule
\end{tabular}
\begin{tabular}{p{0.8\linewidth}p{0.1\linewidth}}
\toprule
1. Please state the relative weight you would assign each of the aspects in the overall context of choosing a job (using the slider).  You are encouraged to use the full range of the scale: - 1. The office & 1 to 8 \\
2. Please state the relative weight you would assign each of the aspects in the overall context of choosing a job (using the slider).  You are encouraged to use the full range of the scale: - 2. The commute & 1 to 8 \\
3. Please state the relative weight you would assign each of the aspects in the overall context of choosing a job (using the slider).  You are encouraged to use the full range of the scale: - 3. The salary & 1 to 8 \\
4. Please state the relative weight you would assign each of the aspects in the overall context of choosing a job (using the slider).  You are encouraged to use the full range of the scale: - 4. The vacation package & 1 to 8 \\
\bottomrule

\end{tabular}%
\caption{Detailed description of the job offer questionnaire that the participants were administered.}
\label{tab:job_offer_qus}
\end{table*}

\section{Prompts}
The zero-shot and few-shot prompts for eliciting the CIS\_Inf scores are shown in Table~\ref{tab:prompts}.
\begin{table*}[]
\small
\begin{tabular}{p{0.1\linewidth}p{0.8\linewidth}}
\toprule
\textbf{Shot} & \textbf{Prompt}\\
\midrule
0-shot & You are an expert social and cognitive psychologist analyzing decision-making patterns from the 2004 study "Construction of Preferences by Constraint Satisfaction".   You are tasked with evaluating how preferences change when participants choose between two job offers with multiple attributes,              in a simulated setting. This experiment measured preferences before and after making a decision,              revealing "coherence shifts" where preferences aligned more closely with the chosen job offer,              occurring both with and without influencing attributes in the job description.              Your goal is to estimate two scores:              (1) the score of a coherence shift towards preferring the chosen job offer, expressed as a value between 0 and 1, where 0 indicates an increased preference for the rejected offer and 1 indicates a strong preference for the chosen offer; and             (2) the score that the decision is influenced by the job descriptions, also on a scale from 0 to 1, where 0 signifies no influence and a rigid preference, and 1 signifies being easily swayed by minor incentives.              Base your assessment on text provided by the user about a recent personal decision that need not be related to the job offer scenario.              Consider the cognitive styles and patterns of decision making evident in their narrative.              Present your findings in this format: "The score of a coherence shift towards the chosen job offer is: \textless{}score\textgreater and the score of being influenced by minor incentives is: \textless{}score\textgreater{},"              with each score ranging between 0 and 1. \\
\midrule
4-shot & You are an expert social and cognitive psychologist analyzing decision-making patterns from the 2004 study "Construction of Preferences by Constraint Satisfaction".              You are tasked with evaluating how preferences change when participants choose between two job offers with multiple attributes,              in a simulated setting. This experiment measured preferences before and after making a decision,              revealing "coherence shifts" where preferences aligned more closely with the chosen job offer,              occurring both with and without influencing attributes in the job description.              Your goal is to estimate two scores based on user-provided text:              (1) the score of a coherence shift towards preferring the chosen job offer, expressed as a value between 0 and 1, where 0 indicates an increased preference for the rejected offer and 1 indicates a strong preference for the chosen offer; and             (2) the score that the decision is influenced by the job descriptions, also on a scale from 0 to 1, where 0 signifies no influence and a rigid preference, and 1 signifies being easily swayed by minor incentives.              Here are four different examples of participants' narratives about recent personal decisions and with a score towards 1 if they had a coherence shift towards the chosen job offer, 0 if coherence shift is towards the rejected offer. Similarly, there is also a score for if being influenced by minor incentives (1 if influenced, 0 if not influenced):                          Example 1:             User's Narrative: "I recently had to decide whether to buy a new car or keep my old one. The new car had better fuel efficiency and more features, but I was attached to my old car due to sentimental reasons. After considering the costs and benefits, I decided to go with the new car."             Output: "The score of a coherence shift towards the chosen job offer is: 0.8 and the score of being influenced by minor incentives is: 0.6."              Example 2:             User's Narrative: "I was choosing between two vacation destinations: a beach resort and a mountain cabin. I love both settings, but ultimately chose the beach resort because it was more affordable and had better amenities."             Output: "The score of a coherence shift towards the chosen job offer is: 0.7 and the score of being influenced by minor incentives is: 0.5."              Example 3:             User's Narrative: "I had to decide whether to take an online course or attend in-person classes for my professional development. The online course was more flexible, but I prefer face-to-face interaction. I chose the online course because it fit better with my schedule."             Output: "The score of a coherence shift towards the chosen job offer is: 0.9 and the score of being influenced by minor incentives is: 0.4."               Similarly, for the following user input text, estimate the scores. Base your assessment on text provided by the user about a recent personal decision that need not be related to the job offer scenario.              Consider the cognitive styles and patterns of decision making evident in their narrative.              Present your findings in this format: "The score of a coherence shift towards the chosen job offer is: \textless{}score\textgreater and the score of being influenced by minor incentives is: \textless{}score\textgreater{},"              with each score ranging between 0 and 1 as a continuous value.\\
\bottomrule
\end{tabular}%
\caption{Zero- and 4-shot prompts for both Llama3.1 and Gemma models.}
\label{tab:prompts}
\end{table*}

\end{document}